# Everybody Likes to Sleep: A Computer-Assisted Comparison of Object Naming Data from 30 Languages

**Alžběta Kučerová**
MCL Chair
University of Passau
Passau, Germany
alzbeta.kucerova@uni-passau.de

**Johann-Mattis List**
MCL Chair
University of Passau
Passau, Germany
mattis.list@uni-passau.de

## Abstract

Object naming – the act of identifying an object with a word or a phrase – is a fundamental skill in interpersonal communication, relevant to many disciplines, such as psycholinguistics, cognitive linguistics, or language and vision research. Object naming datasets, which consist of concept lists with picture pairings, are used to gain insights into how humans access and select names for objects in their surroundings and to study the cognitive processes involved in converting visual stimuli into semantic concepts. Unfortunately, object naming datasets often lack transparency and have a highly idiosyncratic structure. Our study tries to make current object naming data transparent and comparable by using a multilingual, computer-assisted approach that links individual items of object naming lists to unified concepts. Our current sample links 17 object naming datasets that cover 30 languages from 10 different language families. We illustrate how the comparative dataset can be explored by searching for concepts that recur across the majority of datasets and comparing the conceptual spaces of covered object naming datasets with classical basic vocabulary lists from historical linguistics and linguistic typology. Our findings can serve as a basis for enhancing cross-linguistic object naming research and as a guideline for future studies dealing with object naming tasks.

## 1 Introduction

We all choose particular names for objects when using language in daily communication (Silberer et al., 2020). Psychologists have been interested in the topic of object naming since the late 19th century (see e.g. Catell 1886), with object recognition considered a key function of the human brain (Rossion and Pourtois 2004; Hummel 2013; Wardle and Baker 2020). Beyond psychology, researchers have previously also studied its mechanisms in psycholinguistics, computational linguistics, or language and vision research (Silberer et al., 2020).

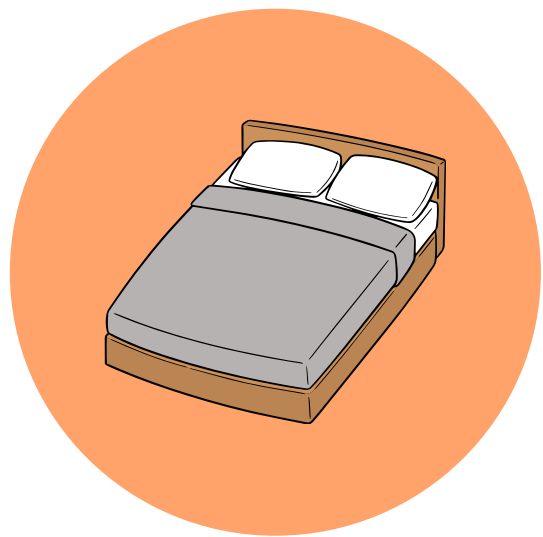

Figure 1: Object naming datasets use picture stimuli similar to this one as a visual cue. The concept BED was used in all of our datasets.

Most scholars to date conduct research monolingually and rarely attempt to explore cross-linguistic perspectives. When they do, they stumble upon issues in replicability, as the concepts they include often do not exist in all languages. Many datasets also include unique, culture-specific items, which pose a bottleneck for low-resource languages or generally any language outside of the Indo-European language family. Time and again, researchers must update or adapt their word lists for each newly tested language, which increases not only the workload but also diminishes the chance of true comparative research (see e.g. Bangalore et al. 2022; Tsaparina et al. 2011; van Dort et al. 2007), and ultimately, the generalizability of any effect across different languages (Blum et al., 2024).

The main idea of this study is to take the first steps towards making current object naming data more transparent and comparable across individual object naming datasets. We address this problem by using a computer-assisted approach to integrate object naming data from various sources into a multilingual dataset. We compare a large multilingual sample of commonly cited object naming datasets (including datasets like Snodgrass and Vanderwart 1980 and Moreno-Martínez and Montoro 2012). In this way, we can obtain insights into commonalities and differences of the semantic space covered by object naming studies. Currently, our sample consists of datasets on object naming from 30 languages and 10 language families. To allow for a systematic comparison, they are linked to norm datasets on concepts in comparative linguistics (Concepticon, https://concepticon.clld.org, List et al. 2025) and can therefore be directly linked to concept norms in psycholinguistics (NoRaRe, https://norare.clld.org, Tjuka et al. 2020).

## 2 Background

### 2.1 Object Naming

When communicating, we often encounter situations in which we want to refer to objects that surround us. While simply pointing to them is possible, we typically use language to refer to these objects, that is, we *name* them. This way, we introduce them into the conversation or direct our conversation partners to them. In disciplines such as psycholinguistics or language and vision research, object naming is an important task that scientists use to gain insights into how humans access and select names for objects in their environment. To study the cognitive processes of converting visual stimuli into semantic concepts, pictures are often used as a visual cue (Ghasisin et al., 2015) – for an example see Figure 1. Scholars in this field of research construct concept lists along with picture-pairings and use them to investigate various research questions such as language production and processing (Tomaschek and Tucker, 2023), aging of the human brain (Connor et al., 2004), aphasia (Paradis, 2011), developmental language disorders (Araújo et al., 2011), or also as a tool for the development and fine-tuning of models in natural language processing (Krishna et al., 2017; Silberer et al., 2020). The extensive use of object naming datasets across a broad range of fields shows that this kind of data might deserve more widespread attention by scholars interested in meaning and semantics.

### 2.2 Concepticons and WordNets

WordNets represent one of the most widespread computer-assisted approaches to meaning and have proven extremely useful in handling various problems in lexical semantics (Rudnicka et al., 2019). When dealing with cross-linguistic resources aiming to compare words and their meanings across multiple languages, however, the use of *Concepticons* (Gaizauskas et al., 1997) presents a straightforward alternative. While a WordNet can be understood as a collection of words in a given language that are assigned to a given number of senses, with the senses being linked to each other by an ontology, a Concepticon seeks to link concepts across multiple languages. The concepts themselves are usually taken from questionnaires or concept lists that are traditionally compiled in historical linguistics, linguistic typology, or language documentation, in order to assemble lexical items for a given language that can be easily compared with lexical items from other languages that express similar or identical meanings.

### 2.3 The CLLD Concepticon

With the CLLD Concepticon (List et al. 2025, https://concepticon.clld.org), a large Concepticon has been compiled that links several hundred concept lists and questionnaires that are used in historical linguistics (Swadesh, 1955), cognitive linguistics (Nicholas et al., 1989), and psycholinguistics (Snodgrass and Vanderwart, 1980) to a common concept space. This space consists of more than 4000 different *concept sets* that are individually defined and linked to individual *elicitation glosses* in individual concept lists. Apart from the data and the web application that can be used to browse through the datasets, the CLLD Concepticon has also established a workflow for the linking of concept lists, which makes use of computer-assisted techniques, providing automatic methods that can be used for the mapping of concept lists to the CLLD Concepticon (List, 2022), and guidelines for individual and collaborative annotation (Tjuka 2020, see also Tjuka et al. 2023).

### 2.4 Norms, Ratings, and Relations

Apart from being useful for the design and investigation of cross-linguistic questionnaires, the CLLD Concepticon can also be used as the basis to increase and enrich all datasets that can be represented in the form of a concept list. An example for this reuse potential is the Database of *Norms, Ratings, and Relations of Words and Concepts* (NoRaRe, Tjuka et al. 2022, https://norare.clld.org/). This database builds on the CLLD Concepticon to harvest various kinds of speech norms, ratings, and additional kinds of semantic metadata (including data from the Open Multilingual Wordnet project, Bond and Foster 2013), that are typically compiled in psycholinguistics and computational linguistics, but barely integrated across individual languages (Tjuka et al., 2020).

| #  | Language   | Dataset                                                      | Reference                                                      |
|----|------------|--------------------------------------------------------------|----------------------------------------------------------------|
| 1  | Arabic     | Boukadi-2015-348, Dunabeitia-2022-500                        | Boukadi et al. 2015, Duñabeitia et al. 2022                    |
| 2  | Hindi      | Ramanujan-2019-158                                           | Ramanujan and Weekes 2019                                      |
| 3  | Russian    | Tsaparina-2011-260, Dunabeitia-2022-500                      | Tsaparina et al. 2011, Duñabeitia et al. 2022                  |
| 4  | Catalan    | Dunabeitia-2022-500                                          | Duñabeitia et al. 2022                                         |
| 5  | German     | Dunabeitia-2018-750, Dunabeitia-2022-500                     | Duñabeitia et al. 2018, Duñabeitia et al. 2022                 |
| 6  | Hungarian  | Dunabeitia-2022-500                                          | Duñabeitia et al. 2022                                         |
| 7  | Slovak     | Dunabeitia-2022-500                                          | Duñabeitia et al. 2022                                         |
| 8  | Czech      | Dunabeitia-2022-500                                          | Duñabeitia et al. 2022                                         |
| 9  | Japanese   | Nishimoto-2005-359                                           | Nishimoto et al. 2005                                          |
| 10 | Turkish    | Raman-2013-260, Dunabeitia-2022-500                          | Raman et al. 2013, Duñabeitia et al. 2022                      |
| 11 | Croatian   | Rogic-2013-346                                               | Rogić et al. 2013                                              |
| 12 | Welsh      | Dunabeitia-2022-500                                          | Duñabeitia et al. 2022                                         |
| 13 | Norwegian  | Dunabeitia-2022-500                                          | Duñabeitia et al. 2022                                         |
| 14 | Basque     | Dunabeitia-2022-500                                          | Duñabeitia et al. 2022                                         |
| 15 | Cantonese  | Zhong-2024-1286                                              | Zhong et al. 2024                                              |
| 16 | Serbian    | Dunabeitia-2022-500                                          | Duñabeitia et al. 2022                                         |
| 17 | Korean     | Hwang-2021-60, Dunabeitia-2022-500                           | Hwang et al. 2021, Duñabeitia et al. 2022                      |
| 18 | Malay      | vanDort-2007-50, Dunabeitia-2022-500                         | van Dort et al. 2007, Duñabeitia et al. 2022                   |
| 19 | Dutch      | Shao-2016-327, Dunabeitia-2018-750, Dunabeitia-2022-500      | Shao and Stiegert 2016, Duñabeitia et al. 2018, Duñabeitia et al. 2022 |
| 20 | Greek      | Dimitropoulou-2009-260                                       | Dimitropoulou et al. 2009                                      |
| 21 | Mandarin   | Liu-2011-435                                                 | Liu et al. 2011                                                |
| 22 | Hebrew     | Dunabeitia-2022-500                                          | Duñabeitia et al. 2022                                         |
| 23 | Spanish    | MorenoMartinez-2012-360, Dunabeitia-2018-750, Dunabeitia-2022-500 | Moreno-Martínez and Montoro 2012, Duñabeitia et al. 2018, Duñabeitia et al. 2022 |
| 24 | Kannada    | Bangalore-2022-180                                           | Bangalore et al. 2022                                          |
| 25 | French     | Dunabeitia-2018-750, Dunabeitia-2018-750                     | Duñabeitia et al. 2018                                         |
| 26 | English    | Snodgrass-1980-260, Dunabeitia-2018-750, Dunabeitia-2022-500 | Snodgrass and Vanderwart 1980, Duñabeitia et al. 2018, Duñabeitia et al. 2022 |
| 27 | Polish     | Dunabeitia-2022-500                                          | Duñabeitia et al. 2022                                         |
| 28 | Finnish    | Dunabeitia-2022-500                                          | Duñabeitia et al. 2022                                         |
| 29 | Italian    | Dunabeitia-2018-750, Dunabeitia-2022-500                     | Duñabeitia et al. 2022                                         |
| 30 | Portuguese | Dunabeitia-2022-500                                          | Duñabeitia et al. 2022                                         |

Table 1: Languages covered in our multilingual sample and their corresponding datasets.

While a full integration of WordNet resources has not been approached so far, neither within the CLLD Concepticon, nor within the NoRaRe database, it is important to emphasize that the lack of integration is not due to the impossibility to integrate WordNets and Concepticons, but rather due to the lack of resources to carry out this task.

## 3 Materials and Methods

### 3.1 Materials

For our analysis, we compiled and investigated 17 distinct object naming datasets of various sizes featuring 30 languages from 10 different language families. A detailed overview can be found in Table 1, with a comprehensive list of all covered languages.

### 3.2 Methods

We mapped all concepts present in our object naming datasets onto existing Concepticon *concept sets* using the standard workflow accessible to everybody using the PyConcepticon library (Forkel et al. 2021, https://pypi.org/project/pyconcepticon, for a detailed description and further requirements see Tjuka 2020). After the initial automatic mapping was finished, we manually checked all entries. In cases, where concepts were matched to incorrect or multiple *concept sets*, we corrected them. Similarly, if no *concept set* was found for certain concepts, but they appeared frequently across our datasets, a new *concept set* was introduced – e.g. PIANO (INSTRUMENT) and TOP (TOY), alongside a short definition. In total, we have introduced 42 new *concept set* to Concepticon, and once this was done, we submitted all of the datasets and new *concept sets* to the public repository of the Concepticon project (https://github.com/concepticon/concepticon-data) and had them reviewed by a team of Concepticon editors for inclusion in the version 3.3 of the CLLD Concepticon (List et al., 2025).

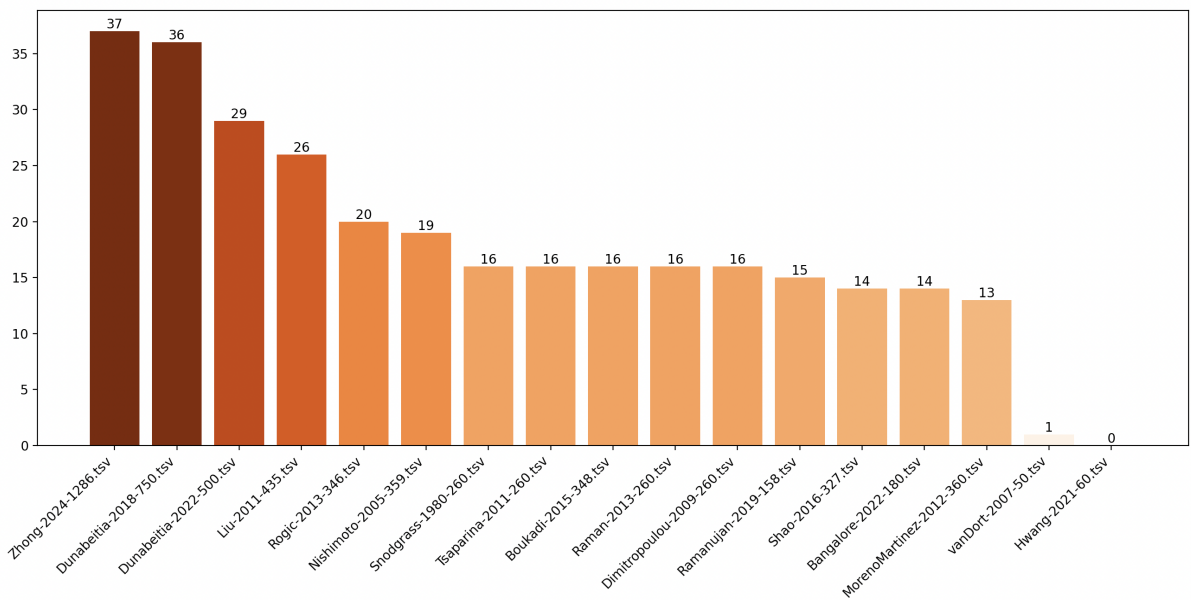

Figure 2: Overlap of our object naming datasets with the Swadesh (1955) list in historical linguistics.

Subsequently, we conducted multiple exploratory analyses of the data. Firstly, we ran a frequency analysis of concepts from all languages, that appeared across all or most datasets. This helped us to identify to which degree concepts differ regarding their suitability for object naming across languages. Secondly, we were interested to see which of the concepts included in our datasets belong to the realm of *basic vocabulary* (Swadesh, 1955). Using the nouns in the 100 item concept list that Swadesh (1955) proposed for the purpose of historical language comparison, we tried to identify which basic concepts recur in object naming studies. Lastly, we assessed all datasets for concepts that only appeared once within the sample in order to get a better understanding regarding the amount of idiosyncratic items in object naming studies.

### 3.3 Implementation

All comparisons, frequency analyses, and visualizations were conducted with the help of Python scripts. All data and code needed to replicate the studies have been curated on CodeBerg (Version 1.0, https://codeberg.org/calc/object-naming-data) and archived with Zenodo (https://doi.org/10.5281/zenodo.14922076).

## 4 Examples

In the following, we illustrate how the linked concept lists can be put to direct use, by assessing the amount of basic vocabulary items in object naming data and by discussing recurring and unique concepts in object naming questionnaires.

### 4.1 Basic Vocabulary

Most object-naming datasets strive to be applicable and relevant across many different cultures and as many languages as possible, to facilitate comparative research. To achieve this, one should elicit words for such concepts that exist in all of the chosen languages. Our analysis, however, revealed that this is not always the case in object naming datasets (for more detail, see Sections 4.2 and 4.3). In historical linguistics, concepts that are expressed regularly by individual words across as many languages as possible have been investigated for a long time in the context of the so-called *basic vocabulary*. Popularized by Morris Swadesh (1909–1967), the idea "that certain parts of the lexicon of human languages are universal, stable over time, and rather resistant to borrowing" (List et al., 2016, 2393) has been fascinating historical linguists for a long time. When dealing with object naming data derived from object naming studies, it may therefore be interesting to test to which degree the basic concepts proposed by historical linguists overlap with the concepts selected by psychologists conducting object namings studies. To answer this question, we compared all our datasets with the basic concepts in Swadesh's list of 100 items (Swadesh, 1955), from which we selected all 54 nouns, given that object naming uses primarily nouns as stimuli. We assessed our datasets for overlap with this list of supposedly stable and universally expressed concepts, looking at the number of basic concepts used across object naming datasets, as well as investigating the particular basic concepts employed by particular datasets. The results show that most datasets employ very few basic concepts. With a decreasing amount of concepts in the dataset in general, the amount of basic concepts also declines. The largest included dataset, Zhong et al. 2024, with a total of 1286 items, contained only 37 of Swadesh's basic concepts. The regularly cited Snodgrass and Vanderwart (1980) dataset, and datasets inspired by it, yielded between 14 and 16 basic concepts from the list of 54 nouns. One dataset in our sample did not include any of Swadesh's basic concepts at all (Hwang et al., 2021). To further elaborate on our analysis, we also looked into the specific basic concepts that the lists include. The results of both analyses are presented in Figures 2 and 3.

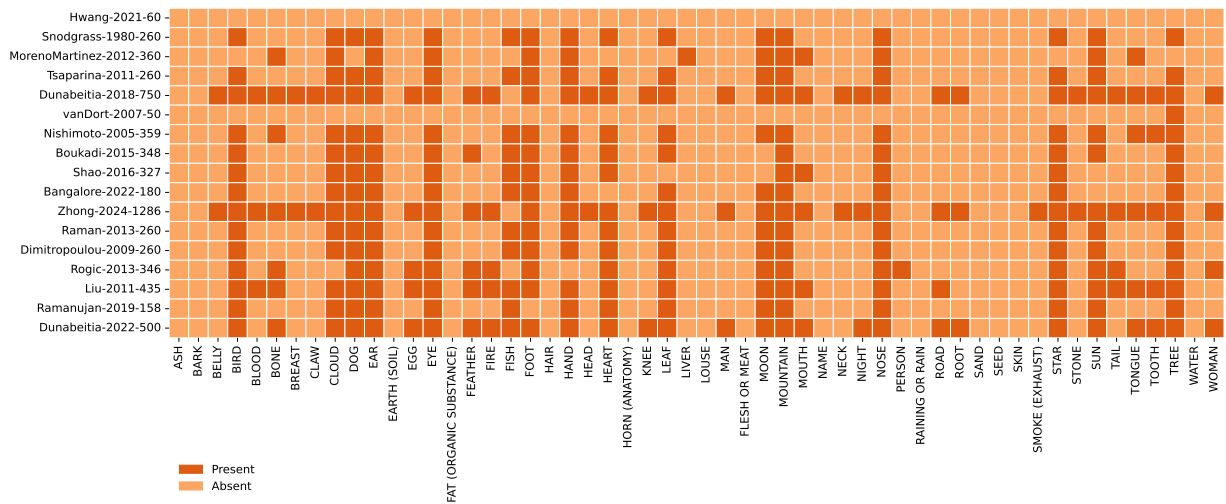

Figure 3: Swadesh concepts in the object-naming datasets (nouns). The y-axis represents the ID of each individual dataset, while the x-axis displays all 54 nouns extracted from the Swadesh (1955) list.

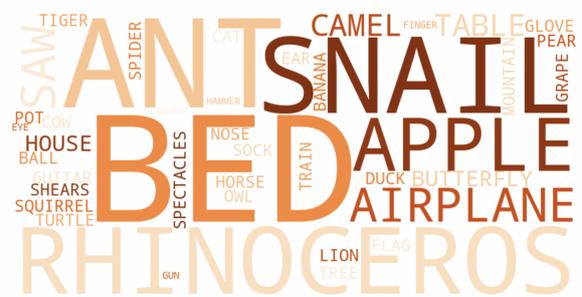

Figure 4: A total of 42 recurring concepts in at least 15 out of 17 datasets - we find BED in all 17 of them.

## 4.2 Recurring Concepts Across Languages

Having mapped such a large number of datasets from as many as 30 languages, we sought to determine whether there was any common ground or standard in the authors' choice of concepts. We hypothesized that this would be the case. Early investigations have shown that this overlap would, however, be much smaller than expected. At the end, we compared all of our 17 datasets with each other and discovered that the overlap consists of exactly one word: BED – hence our title. We therefore decided to estimate two further benchmarks for the assessment of shared concepts, specifically words that appear in at least 16 out of the 17 datasets and words that appear in at least 15 out of the 17 datasets. This has proven rather useful and the result of this can be seen in Figure 4.

With the first benchmark, our lists had an overlap of 11 concepts: ANT, SAW, AIRPLANE, TABLE, CAMEL, BUTTERFLY, HOUSE, SNAIL, RHINOCEROS, APPLE, and BED, but with the second, slightly larger one, this number increased to 42 items. We were interested to see, which datasets were the ones that did not include certain concepts. An overview is shown in Table 2, for the numbers of datasets, please refer to Table 1. One can observe that it is mostly two datasets that often lack certain concepts. These are Hwang et al. (2021) and van Dort et al. (2007), two monolingual datasets on Korean and Malay, respectively. Both of these languages are, however, included within other datasets as well. The reason behind them not featuring these concepts is likely their size. In comparison to the other datasets, they only feature 60 and 50 items and have therefore *fewer spaces to fill*. In the case of other larger datasets, such as e.g. Ramanujan and Weekes, 2019 (Hindi) and Bangalore et al., 2022 (Kannada), missing items might not be featured because they are not native to the area, e.g. GUITAR. This is something a researcher attempting to create a multilingual dataset should always keep in mind.

| Concept | Absent | Concept | Absent | Concept | Absent | Concept | Absent | Concept | Absent | Concept | Absent |
|---|---|---|---|---|---|---|---|---|---|---|---|
| AIRPLANE | 9 | CAT | 9, 8 | GLOVE | 9, 10 | MOUNTAIN | 9, 8 | SNAIL | 8 | TRAIN | 9, 8 |
| ANT | 9 | COW | 9, 8 | GRAPE | 9, 13 | NOSE | 9, 8 | TREE | 14, 8 | TURTLE | 9, 8 |
| APPLE | 9 | DUCK | 9, 8 | GUITAR | 2, 15 | OWL | 9, 8 | SOCK | 9, 8 | | |
| BALL | 9, 8 | EAR | 9, 8 | GUN | 9, 8 | PEAR | 9, 8 | SPECTACLES | 9, 14 | | |
| BANANA | 9, 8 | ELEPHANT | 9, 8 | HAMMER | 9, 8 | POT | 9, 8 | SPIDER | 9, 8 | | |
| **BED** | – | EYE | 9, 8 | HORSE | 9, 8 | RHINOCEROS | 6 | SQUIRREL | 10, 14 | | |
| BUTTERFLY | 9 | FINGER | 9, 8 | HOUSE | 8 | SAW | 8 | TABLE | 9 | | |
| CAMEL | 8 | FLAG | 14, 8 | LION | 9, 8 | SHEARS | 15, 14 | TIGER | 9, 15 | | |

Table 2: Recurring concepts in at least 15 out of 17 datasets with a specification of the dataset, in which they are absent. BED is the only concept that is present in all of them. For number references, see Table 1.

### 4.3 Unique Concepts

Not all *concept sets* recur in object naming datasets. We find that many researchers include highly specific or niche items, which can act as a constraint when it comes to later application of a dataset in low-resource languages, or generally any language that falls outside the Indo-European language family. Examples of such concepts include items like: DANDELION, EUCALYPTUS, POLAR BEAR, or culturally specific items, such as CHURRERA or FOOTBALL HELMET. On the other hand, some authors incorporate otherwise underrepresented concepts from Swadesh (1955), such as LIVER and SMOKE (EXHAUST), which would be advantageous in cross-linguistic, comparative studies.

## 5 Conclusion

Object naming is a growing and relevant field in many scientific disciplines. New datasets are produced, adapted, or updated every year. Through a computer-assisted analysis of 17 datasets from 30 languages and 10 language families, our study demonstrates that transparency and cross-linguistic comparability of object naming data can be achieved, for example, by mapping existing concept lists onto *concept sets* in Concepticon. By applying computational methods, we have assessed, compared, and evaluated an extensive multilingual dataset, providing insights into the general strategies and patterns, as well as the covered semantic space. We have shown that object naming datasets employ some basic concepts as per Swadesh (1955), even though the number is smaller than expected. Additionally, we found that although object naming datasets differ greatly, there appear to be several recurring concepts – 42 in total. Lastly, we showed that some concepts appear exclusively in certain datasets and often introduce overly specific, niche or cultural items that pose a bottleneck for low-resource languages. Because most of the covered object naming datasets include valuable psycholinguistic norms, they will be linked to NoRaRe in the future. Our study can serve as a basis for enhancing comparative and transparent cross-linguistic object naming research and as a guideline for future studies in the field.

### Limitations

Our study by now follows a concept-based approach, using the CLLD Concepticon. However, in the future, our findings can be extended using additional semantic technologies. Thus, as suggested by one of our reviewers, the integration with the Open Multilingual WordNet should be explored, as it might provide new insights and prove useful for resources that could build on our object naming data. We would also like to explore the integration with Interlingual Indices (Bond et al., 2016, Vossen et al., 1999), as a WordNet technology that offers services similar to a Concepticon.

## Supplementary Material

Data and code have been curated on CodeBerg (Version 1.0, https://codeberg.org/calc/object-naming-data) and archived with Zenodo (https://doi.org/10.5281/zenodo.14922076).

## Acknowledgements

This project was supported by the ERC Consolidator Grant ProduSemy (PI Johann-Mattis List, Grant No. 101044282, see https://doi.org/10.3030/101044282). Views and opinions expressed are, however, those of the authors only and do not necessarily reflect those of the European Union or the European Research Council Executive Agency (nor any other funding agencies involved). Neither the European Union nor the granting authority can be held responsible for them. We would like to thank the anonymous reviewers for their helpful comments. We also thank Frederic Blum and Annika Tjuka for their assistance with the collaborative review of our data for the inclusion in the Concepticon project. We further express our gratitude to all people who share their data openly, so we can use them in our research.